# Evaluating Influence Diagrams using LIMIDs

Dennis Nilsson   Steffen L. Lauritzen
Department of Mathematical Sciences
Aalborg University
Fredrik Bajers Vej 7E
9220 Aalborg
Denmark

## Abstract

We present a new approach to the solution of decision problems formulated as influence diagrams. The approach converts the influence diagram into a simpler structure, the **LI**mited **M**emory **I**nfluence **D**iagram (LIMID), where only the requisite information for the computation of optimal policies is depicted. Because the requisite information is explicitly represented in the diagram, the evaluation procedure can take advantage of it. In this paper we show how to convert an influence diagram to a LIMID and describe the procedure for finding an optimal strategy. Our approach can yield significant savings of memory and computational time when compared to traditional methods.

## 1 INTRODUCTION

Influence Diagrams (IDs) were introduced by Howard and Matheson (1981) as a compact representation of decision problems. Since then, various authors have attempted to formalize their approach and develop algorithms for evaluating IDs.

Olmsted (1983) and Shachter (1986) initiated research in this direction. Their methods operate directly on the ID and consist of eliminating nodes from the diagram through a series of value preserving transformations. During the transformations the policies for the decisions are computed. Later Shachter and Ndilikilikesha (1993) and Ndilikilikesha (1994) proposed a similar, but more efficient approach.

Other algorithms evaluate IDs by converting them into different structures. Cooper (1988) described an approach where the evaluation of IDs is transformed into inference problems for Bayesian networks. Several improvements of this method were later proposed by Shachter and Peot (1992) and Zhang (1998). Shenoy (1992) presented a method where the ID is converted into a valuation network, and the optimal strategy is computed through the removal of nodes from this diagram by fusing the valuations bearing on the node to be removed. Jensen *et al.* (1994) compiled the ID into a secondary structure, the strong junction tree, and solved the decision problem by the passage of messages towards the root of the tree.

Our work relies on a property that has already been stressed by Shachter (1998, 1999), and Nielsen and Jensen (1999). Namely that in decision problems represented as IDs there may be information which is not requisite for computing the policies. Going further, we transform the ID into a similar, but simpler, structure termed **LI**mited **M**emory **I**nfluence **D**iagram (LIMID) where the requisite information is explicitly depicted, and present a simple algorithm for finding the optimal strategy using this reduced structure. This can result in significant gains in efficiency compared to traditional methods for solving IDs.

Section 2 gives a basic description of LIMIDs as developed in Lauritzen and Nilsson (1999). For proofs not given in the present paper, the reader is referred to this source.

## 2 LIMIDS

LIMIDs are represented by directed acyclic graphs (DAGs) with three types of nodes. *Chance nodes*, shown as circles, represent random variables. *Decision nodes*, shown as squares, represent choices or actions available to the decision maker. Finally, *value nodes*, shown as diamonds, represent local utility functions. The arcs in a LIMID have a different meaning based on their target. Arcs pointing to utility or chance nodes represent probabilistic or functional dependence. Arcs into decision nodes indicate which variables are known to the decision maker at the time of decision. Thus they in particular imply time precedence.

In contrast with traditional IDs, the LIMID can represent decision problems that violates the assumption of *no forgetting* saying that variables known at the time of one decision must also be known when all later decisions are made.

The following fictitious decision problem borrowed from Lauritzen and Nilsson (1999) illustrates a typical decision situation which is well described by a LIMID.

> A pig breeder is growing pigs for a period of four months and subsequently selling them. During this period the pig may or may not develop a certain disease. If the pig has the disease at the time when it must be sold, the pig must be sold for slaughtering. On the other hand, if it is disease free, its expected market price as a breeding animal is higher. Once a month, a veterinary doctor sees the pig and makes a test for presence of the disease. The test result is not fully reliable and will only reveal the true condition ($h_i$) of the pig with a certain probability. Based on the test result ($t_i$), the doctor decides whether treating the pig for the disease ($d_i$).

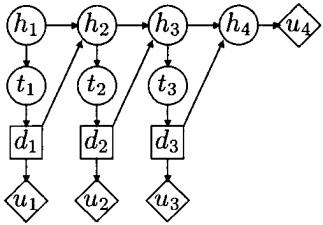

The diagram above represents the LIMID corresponding to the situation where the pig breeder does not keep individual records for his pigs and has to make his decision knowing only the given test result. The memory has been limited to the extreme of only remembering the present. In the LIMID, the utility nodes $u_1, u_2, u_3$ represent the potential treatment costs, whereas $u_4$ is the (expected) market price of the pig as determined by its health at the fourth month.

### 2.1 SPECIFICATION OF LIMIDS

Suppose we are given a LIMID $\mathcal{L}$ with decision nodes $\Delta$ and chance nodes $\Gamma$. We let $V = \Delta \cup \Gamma$. The set of value nodes is denoted $\Upsilon$.

For a node $n$ we let $\mathrm{pa}(n)$ denote its parents. Each node $n \in V$ is associated with a variable which we likewise denote by $n$, that takes a value in a finite set $\mathcal{X}_n$. For $W \subseteq V$ we write $\mathcal{X}_W = \times_{n \in W} \mathcal{X}_n$. Typical elements in $\mathcal{X}_W$ are denoted by lower case values such as $x_W$, abbreviating $x_V$ to $x$.

Associated with every chance node $r$ (connoting random variable) is a non-negative function $p_r$ on $\mathcal{X}_r \times \mathcal{X}_{\mathrm{pa}(r)}$ such that

$$p_r(x_r \mid x_{\mathrm{pa}(r)}) \geq 0 \quad \sum_{x_r} p_r(x_r \mid x_{\mathrm{pa}(r)}) = 1, \quad (1)$$

where the sum is over $\mathcal{X}_r$. The term $p_r$ does not in general correspond to a true conditional distribution but rather a family of probability distributions for $r$ parametrized by the states of $\mathrm{pa}(r)$.

Each value node $u \in \Upsilon$ is associated with a real function $U_u$ defined on $\mathcal{X}_{\mathrm{pa}(u)}$.

### 2.2 POLICIES AND STRATEGIES

A policy for decision node $d$ can be regarded as a prescription of alternatives in $\mathcal{X}_d$ for each possible observation in $\mathcal{X}_{\mathrm{pa}(d)}$. To allow for the possibility of randomizing between alternatives, we formally define a policy as follows. A *policy* $\delta_d$ for $d$ is a non-negative function on $\mathcal{X}_d \times \mathcal{X}_{\mathrm{pa}(d)}$ which indicates a probability distribution over alternative choices for each possible value of $\mathrm{pa}(d)$. They must also satisfy the relation (1) as above. A *strategy* is a collection of policies $\{\delta_d : d \in \Delta\}$, one for each decision.

A strategy $q = \{\delta_d : d \in \Delta\}$ determines a joint distribution of all the variables in $V$ as

$$f_q = \prod_{r \in \Gamma} p_r \prod_{d \in \Delta} \delta_d, \quad (2)$$

and $p_r$ and $\delta_d$ are indeed true conditional distributions w.r.t. $f_q$.

The expected utility of the strategy $q$ is given by

$$\mathrm{EU}(q) = \sum_x f_q(x) U(x),$$

where $U = \sum_{u \in \Upsilon} U_u$ is the total utility. We are searching for an optimal strategy $\hat{q}$ satisfying

$$\mathrm{EU}(\hat{q}) \geq \mathrm{EU}(q) \text{ for all strategies } q.$$

Such an optimal strategy is termed a *global maximum strategy* in Lauritzen and Nilsson (1999).

### 2.3 SOLUBLE LIMIDS

The complexity of finding optimal strategies within LIMIDs is in general prohibitive. This task, however, becomes feasible for LIMIDs that have a certain structure. For that reason they are termed *soluble*. In this section we formally define soluble LIMIDs and present a simple and efficient algorithm for evaluating them.

For a strategy $q = \{\delta_d : d \in \Delta\}$ and any $d_0 \in \Delta$ we let

$$q_{-d_0} = q \setminus \{\delta_{d_0}\}$$

be the partially specified strategy obtained by retracting the policy at $d_0$.

A *local maximum policy* for a strategy $q$ at $d_0$, is a policy $\delta'_{d_0}$ which satisfies

$$\mathrm{EU}\left(\{\delta'_{d_0}\} \cup q_{-d_0}\right) = \sup_{\delta_{d_0}} \mathrm{EU}\left(\{\delta_{d_0}\} \cup q_{-d_0}\right)$$

So, $\delta'_{d_0}$ is a local maximum policy for $q$ at $d_0$ if and only if the expected utility does not increase by changing the policy $\delta'_{d_0}$ given the other policies are as in $q$. The following lemma gives a method to find a local maximum policy. Here, $f_{q_{-d}}$ is defined through (2) and the partial strategy obtained from $q$ by retracting $\delta_d$. Letting the *family of $n$* be defined by $\mathrm{fa}(n) = \mathrm{pa}(n) \cup \{n\}$ we now have

**Lemma 1** *A policy $\tilde{\delta}_d$ is a local maximum policy for a strategy $q$ at $d$ if and only if for all $x_{\mathrm{fa}(d)}$ with $\tilde{\delta}_d(x_d \mid x_{\mathrm{pa}(d)}) > 0$ we have*

$$x_d = \arg\max_{z_d} \sum_{x_{V \setminus \mathrm{fa}(d)}} f_{q_{-d}}(x_{V \setminus d}, z_d) U(x_{V \setminus d}, z_d).$$

As we shall see in Theorem 1, an important instance of Lemma 1 is when the strategy $q$ is the *uniform strategy*. Here the uniform strategy $\bar{q}$ is defined as the strategy $\bar{q} = \{\bar{\delta}_d : d \in \Delta\}$, where

$$\bar{\delta}_d(x_d \mid x_{\mathrm{pa}(d)}) = 1/|\mathcal{X}_d|.$$

Letting

$$f = \prod_{r \in \Gamma} p_r, \qquad (3)$$

we now have the following special case of Lemma 1.

**Corollary 1** *A policy $\tilde{\delta}_d$ is a local maximum policy for the uniform strategy at $d$ if and only if for all $x_{\mathrm{fa}(d)}$ with $\tilde{\delta}_d(x_d \mid x_{\mathrm{pa}(d)}) > 0$ we have*

$$x_d = \arg\max_{z_d} \sum_{x_{V \setminus \mathrm{fa}(d)}} f(x_{V \setminus d}, z_d) U(x_{V \setminus d}, z_d).$$

**Proof:** For the uniform strategy $\bar{q}$ we have from (2) and (3) that $f_{\bar{q}_{-d}} \propto f$. Now the corollary follows from Lemma 1. •

An *optimum policy* for $d_0$ in the LIMID $\mathcal{L}$ is a policy which is a local maximum policy at $d_0$ for all strategies $q$ in $\mathcal{L}$. Evidently some decision nodes may not have an optimum policy. However, in the following we present a method for (graphically) identifying decision nodes that have an optimum policy. For this purpose we let the symbolic expression

$$A \perp_{\mathcal{L}} B \mid S$$

denote that $A$ and $B$ are d-separated by $S$ in the DAG formed by all the nodes in the LIMID $\mathcal{L}$, i.e. including the utility nodes.

For a node $n$ we let $\mathrm{de}(n)$ denote the descendants of $n$. We say that a decision node $d_0$ is *extremal* in the LIMID $\mathcal{L}$ if

$$u \perp_{\mathcal{L}} \left(\cup \{\mathrm{fa}(d) : d \in \Delta \setminus \{d_0\}\}\right) \Big| \mathrm{fa}(d_0)$$

for every utility node $u \in \mathrm{de}(d_0)$.

Theorem 1 establishes the connection between optimum policies and extremal decision nodes.

**Theorem 1** *If decision node $d$ is extremal in the LIMID $\mathcal{L}$, then*

- *$d$ has an optimum policy;*
- *any local maximum policy for the uniform strategy at $d$ is an optimum policy for $d$.*

Suppose decision node $d$ is extremal in the LIMID $\mathcal{L}$. Then Theorem 1 ensures that $d$ has an optimum policy $\tilde{\delta}_d$. We can now implement $\tilde{\delta}_d$ by converting $d$ into a chance node with $\tilde{\delta}_d$ as the associated conditional probability distribution to obtain a new LIMID $\mathcal{L}^*$. It is easily seen that every optimal strategy $q^*$ for $\mathcal{L}^*$ then generates an optimal strategy for $\mathcal{L}$ as $q = q^* \cup \{\tilde{\delta}_d\}$. Thus, if $\mathcal{L}^*$ again has an extremal decision node, we can yet again find an optimum policy and convert $\mathcal{L}^*$ as above. If the process can continue until all decision nodes have become chance nodes, we have clearly obtained an optimal strategy for $\mathcal{L}$.

We thus define an *exact solution ordering* $d_1, \ldots, d_k$ of the decision nodes in $\mathcal{L}$ as an ordering with the property that for all $i$, $d_i$ is extremal in the LIMID where $d_{i+1}, \ldots, d_k$ have been converted into chance nodes. A LIMID $\mathcal{L}$ is said to be *soluble* if it admits an exact solution ordering.

Accordingly, computing an optimal strategy for a soluble LIMID $\mathcal{L}$ can be done using the following routine:

**Algorithm** SINGLE POLICY UPDATING

**Input:** A soluble LIMID $\mathcal{L}$ with exact solution ordering $d_1, \ldots, d_k$.

   For $i = k, \ldots, 1$ do:

   1. Compute an optimum policy $\tilde{\delta}_{d_i}$ for $d_i$;
   2. Convert $d_i$ into a chance node with $\tilde{\delta}_{d_i}$ as its associated conditional probability function.

**Return:** The policies $\{\tilde{\delta}_{d_k}, \ldots, \tilde{\delta}_{d_1}\}$.

Note that the policy $\tilde{\delta}_{d_i}$ computed in step 1 is only optimum for $d_i$ in the LIMID where decision nodes $d_{i+1}, \ldots, d_k$ are converted into chance nodes. The algorithm is well-defined since, as described above, the solubility of $\mathcal{L}$ guarantees that it is always possible to compute an optimum policy for $d_i$ in step 1. Thus the collection $\{\tilde{\delta}_{d_k}, \ldots, \tilde{\delta}_{d_1}\}$ constitutes an optimal strategy for $\mathcal{L}$.

## 3  EVALUATING INFLUENCE DIAGRAMS USING LIMIDS

Suppose we are given a decision problem represented by an ID and wish to evaluate it using the algorithm SINGLE POLICY UPDATING. Then one first needs to transform the ID into an 'equivalent' LIMID. This is an easy task: The ID requires a linear temporal order on the decision nodes and, in addition it assumes 'no forgetting', i.e. all variables known at the time of one decision are assumed to be known when subsequent decisions are made. Thus, for an ID with decision nodes $d_1, \ldots, d_k$ (where their index indicate the order of the decisions), the no forgetting assumption can be made explicit by drawing arcs from $\text{fa}(d_j)$ into $d_i$ for all $i$ and for all $j < i$. We call the diagram produced in this way the *LIMID version of the ID*. In Fig. 1-2, an ID and its LIMID version are shown. Now we have

**Theorem 2** *The LIMID version of an ID is soluble.*

**Proof:** Suppose we are given an ID with decision nodes $d_1, \ldots, d_k$. For the LIMID version $\mathcal{L}$ of the ID we have

$$\left(\cup\{\text{fa}(d_j) : j < i\}\right) \subseteq \text{pa}(d_i) \subset \text{fa}(d_i)$$

for all $i$, so $d_i$ is clearly extremal after making $d_{i+1}, \ldots, d_k$ into chance nodes. Thus $\mathcal{L}$ is soluble with exact solution ordering $d_1, \ldots, d_k$.  •

### 3.1  REDUCING SOLUBLE LIMIDS

Starting from a soluble LIMID $\mathcal{L}$ we now present a method for identifying parents of decision nodes that are non-requisite for the computation of optimum policies. Similar methods for IDs have been produced by Nielsen and Jensen (1999) and Shachter (1999) and when a LIMID is representing an ID their mehod identifies the same requisite parents as ours, but the subsequent use of SINGLE POLICY UPDATING exploits this reduction to obtain lower complexity of the computations.

As for IDs the key to simplification of computational problems for LIMIDs is the notion of irrelevance as expressed through the notion of d-separation (Pearl 1986). We say that a node $n \in \text{pa}(d)$ in $\mathcal{L}$ is *non-requisite* for $d$ if

$$u \perp_{\mathcal{L}} n \mid (\text{fa}(d) \setminus \{n\}),$$

for every utility node $u \in \text{de}(d)$. If the above condition is not satisfied, then $n$ is said to be *requisite* for $d$.

A *reduction* of $\mathcal{L}$ is a LIMID obtained by successive removals of arcs from non-requisite parents of decision nodes. It can be shown that any LIMID $\mathcal{L}$ has a unique *minimal reduction*, denoted $\mathcal{L}_{min}$, obtained by reducing $\mathcal{L}$ as much as possible. Thus in $\mathcal{L}_{min}$ all parents of decision nodes are requisite (cf. Theorem 4 in Lauritzen and Nilsson (1999)).

Reducing a soluble LIMID to its minimal reduction can be done by applying the following routine. Note that the algorithm runs in time $O(k(\text{graph size}))$.

**Algorithm Reducing Soluble LIMIDs**

**Input:** A soluble LIMID with exact solution ordering $d_1, \ldots, d_k$.

**For $i = k, \ldots, 1$ do:** Remove arcs from non-requisite parents of decision node $d_i$.

Note that in the above algorithm the decision nodes are visited in the reverse order starting from $d_k$. This ordering is important: If we chose some other ordering there is no guarantee that the reduced LIMID is minimal. For a discussion of this issue the reader is referred to Lauritzen and Nilsson (1999).

Fortunately, the maximum expected utility is preserved under reduction, i.e. if $\mathcal{L}'$ is a reduction of $\mathcal{L}$, then the optimal strategy in $\mathcal{L}'$ and the optimal strategy in $\mathcal{L}$ have the same expected utility. In addition, solubility is preserved under reduction, i.e. any reduction of a soluble LIMID $\mathcal{L}$ is itself soluble. The reader interested in the details and proofs is referred the above source; here we shall use the following theorem.

**Theorem 3** *If the LIMID $\mathcal{L}$ is soluble, then*

1. *its minimal reduction $\mathcal{L}_{min}$ is soluble;*

2. *any optimal strategy for $\mathcal{L}_{min}$ is an optimal strategy for $\mathcal{L}$.*

**Example 1** Regard the ID in Fig. 1, and its LIMID version depicted in Fig 2. The latter diagram is the starting point for reducing the decision problem using Procedure 1:

First one notes that $u_2$ and $u_4$ are the only utility nodes that are descendants of $d_4$. Furthermore

$$\{d_1, d_3\} \perp_{\mathcal{L}} \{u_2, u_4\} \mid (\text{fa}(d_4) \setminus \{d_1, d_3\}),$$

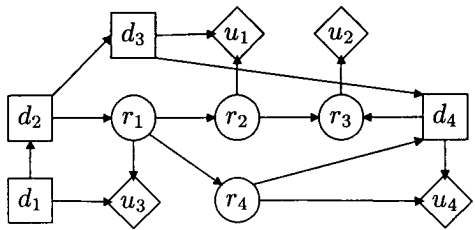

Figure 1: An influence diagram.

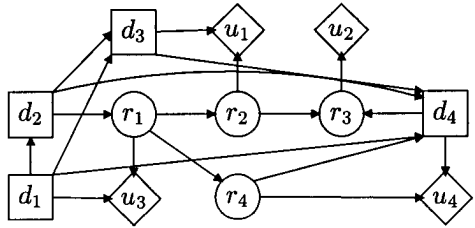

Figure 2: The LIMID version of the ID in Fig. 1.

so $d_1$ and $d_3$ are non-requisite parents of $d_4$. So the arcs from $d_1$ and $d_3$ into $d_4$ are removed and we let $\mathcal{L}_1$ denote the reduced LIMID. Now one notes that in $\mathcal{L}_1$, $u_1$ is the only utility node that is a descendant of $d_3$ and since

$$d_1 \perp_{\mathcal{L}_1} u_1 \mid (\mathrm{fa}(d_3) \setminus \{d_1\}),$$

$d_1$ is non-requisite for $d_3$ and the arc from $d_1$ into $d_3$ is removed. In the reduced LIMID it can be seen that $d_1$ (which is the only parent of $d_2$) is requisite for $d_2$. Finally, $d_1$ has no parents so no further reduction is possible and therefore the reduced LIMID, shown in Fig. 3, is minimal.

## 3.2 CONSTRUCTION OF JUNCTION TREES

As we shall see, computing optimum policies for the decisions during SINGLE POLICY UPDATING can be done by message passing in a so-called junction tree. In the present section we describe how to compile a soluble LIMID into the junction tree. Clearly it is always advantageous to start with a minimal LIMID:

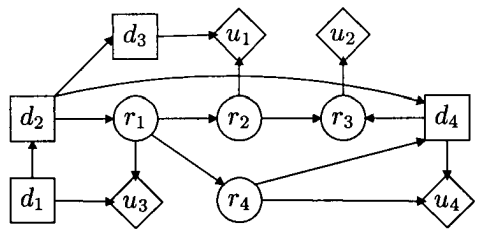

Figure 3: The minimal reduction of the LIMID in Fig. 2

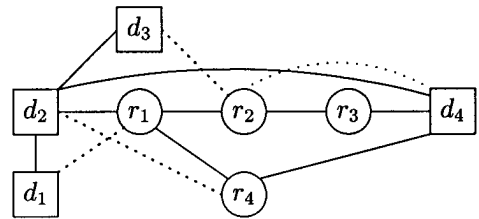

Figure 4: The moralized graph of the LIMID in Fig. 3

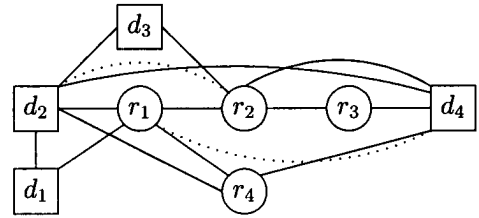

Figure 5: The triangulated graph of the moral graph in Fig. 4. The elimination order used in the triangulation process was $d_1, r_3, d_3, r_4, d_2, r_1, r_2, d_4$.

while not affecting the correctness of the algorithm, the arcs from non-requisite parents introduce unnecessary computations.

The transformation from a LIMID $\mathcal{L}$ to a junction tree starts by adding undirected edges between all nodes with a common child (including children that are decision nodes). Then we drop the directions on all arcs and remove all value nodes to obtain the *moral graph*. Next, edges are added to the moral graph to form a triangulated graph and the cliques are subsequently organized into a junction tree. This can be done in a number of ways; we refer to Cowell *et al.* (1999) for details. It is important to note that, in contrast with the local computation method described by Jensen *et al.* (1994) the triangulation does not need to respect any specific partial ordering of the nodes, but the triangulation can simply be chosen to minimize the computational costs, for example as described in Kjærulff (1992).

**Example 2** Fig. 4 shows the moral graph of the minimal LIMID in Fig. 3, and Fig. 5 displays the triangulation of the moral graph. The elimination order used in the triangulation process is chosen to minimize the size of the cliques; in particular the ordering does not respect the partial ordering of the nodes in the minimal LIMID. The junction tree for the triangulated graph is given in Fig. 6.

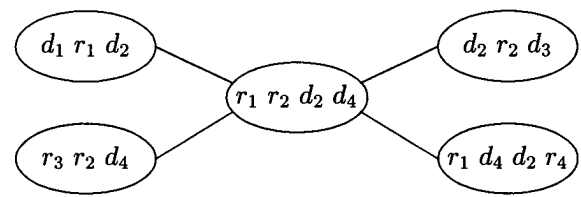

Figure 6: The junction tree of the triangulated graph in Fig. 5.

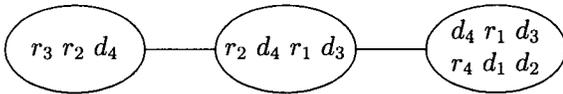

Figure 7: The strong junction tree of the original ID represented in Fig. 1. The rightmost clique is the strong root.

For comparison we have shown the strong junction tree in Fig. 7. Even though the latter has fewer cliques than our junction tree, the largest clique in the strong junction tree contains six variables whereas the largest clique in our junction tree only contains four variables. This is important since the largest clique in the junction tree mainly determines the complexity of message passing in the junction tree.

### 3.3 LIMID POTENTIALS

In our junction tree we represent the quantitative elements of a LIMID through entities called *LIMID-potentials*, or just *potentials* for short.

Let $W \subseteq V$. A potential on $W$ is a pair $\pi_W = (p_W, u_W)$ where

- $p_W$ is a non-negative real function defined on $\mathcal{X}_W$;
- $u_W$ is a real function defined on $\mathcal{X}_W$.

So a potential consists of two parts where the first part $p_W$ is called the *probability part* and the second part $u_W$ is called the *utility part*. A potential is called *vacuous* if its probability part is equal to unity and its utility part is equal to zero. To evaluate the decision problem in terms of potentials we define two basic operations of *combination* and *marginalization*. This notion of operations is similar to what is used in Shenoy (1992), Jensen et al. (1994), and Cowell et al. (1999).

The combination of two potentials $\pi_{W_1} = (p_{W_1}, u_{W_1})$ and $\pi_{W_2} = (p_{W_2}, u_{W_2})$, denoted $\pi_{W_1} \otimes \pi_{W_2}$, is the potential on $W_1 \cup W_2$ given by

$$\pi_{W_1} \otimes \pi_{W_2} = (p_{W_1} p_{W_2}, u_{W_1} + u_{W_2}).$$

The marginalization of the potential $\pi_W = (p_W, u_W)$ onto $W_1 \subseteq W$, denoted $\pi_W^{\downarrow W_1}$ is the potential on $W_1$ given by

$$\pi_W^{\downarrow W_1} = \left( \sum_{W \setminus W_1} p_W, \frac{\sum_{W \setminus W_1} p_W u_W}{\sum_{W \setminus W_1} p_W} \right).$$

Here we have used the convention that $0/0 = 0$.

Two potentials $\pi_W^1 = (p_W^1, u_W^1)$ and $\pi_W^2 = (p_W^2, u_W^2)$ are considered equal, and we write $\pi_W^1 = \pi_W^2$ if for all $x_W$ we have

- $p_W^1(x_W) = p_W^2(x_W)$ and
- $u_W^1(x_W) = u_W^2(x_W)$ whenever $p_W^1(x_W) > 0$.

This identification of two potentials is needed to prove that marginalization and combination satisfy the axioms of Shenoy and Shafer (1990) (cf. Lemma 2–4 in Lauritzen and Nilsson (1999)). This in turn establishes the correctness of the message passing scheme presented in Section 3.5.

### 3.4 INITIALIZATION

To initialize the junction tree $\mathcal{T}$, one assigns a vacuous potential to each clique $C \in \mathcal{C}$. Then for each chance node $r$ in the LIMID $\mathcal{L}$ one multiplies the conditional probability function $p_r$ onto the probability part of any clique containing fa($r$). When this has been done, one takes each value node $u$, and adds the local utility function $U_u$ to the utility part of the potential of any clique containing pa($u$). The moralization process has ensured the existence of such cliques.

Let $\pi_C = (p_C, u_C)$ be the potential on clique $C$ after initialization. The *joint potential* $\pi_V$ on $\mathcal{T}$ is equal to the combination of all potentials and satisfies:

$$\pi_V = \otimes_{C \in \mathcal{C}} \pi_C = \left( \prod_{r \in \Gamma} p_r, \sum_{u \in \Upsilon} U_u \right) = (f, U), \quad (4)$$

where $f$ is defined in (3) and $U$ is the total utility.

### 3.5 PASSAGE OF MESSAGES

Let $\{\pi_C : C \in \mathcal{C}\}$ be a collection of potentials on the junction tree $\mathcal{T}$, and let $\pi_V = \otimes \{\pi_C : C \in \mathcal{C}\}$ be the joint potential on $\mathcal{T}$. Suppose we wish to find the marginal $\pi_V^{\downarrow R}$ for some clique $R \in \mathcal{C}$. To achieve our purpose we present a propagation scheme where messages are passed via a mailbox placed on each edge of the junction tree. If the edge connects $C_1$ and $C_2$, the mailbox can hold messages in the form of potentials on $C_1 \cap C_2$. So when a message is passed from $C_1$ to $C_2$ or vice versa, the message is inserted into the mailbox.

Imagine for the moment that we direct all the edges in $\mathcal{T}$ towards the 'root-clique' $R$. Then each clique passes a message to its child after having received messages from all its other neighbours. The structure of a *message* $\pi_{C_1 \to C_2}$ from clique $C_1$ to its neighbour $C_2$ is given by

$$\pi_{C_1 \to C_2} = \left( \pi_{C_1} \otimes \left( \otimes_{C \in \mathrm{ne}(C_1) \setminus \{C_2\}} \pi_{C \to C_1} \right) \right)^{\downarrow C_2},$$

where ne$(C_1)$ are the neighbours of $C_1$ in $\mathcal{T}$.

In words, the message which $C_1$ sends to its neighbour $C_2$ is the combination of all the messages that $C_1$ receives from its other neighbours together with its own potential, suitably marginalized.

The following result follows from the fact that the two mappings, combination ($\otimes$) and marginalization ($\downarrow$) obey the Shafer–Shenoy axioms.

**Theorem 4** *Suppose we start with a joint potential $\pi_V$ on a junction tree $\mathcal{T}$ with cliques $\mathcal{C}$, and pass messages towards a 'root-clique' $R$ as described above. When $R$ has received a message from each of its neighbours* ne$(R)$, *the combination of all messages with its own potential is equal to the marginalization of $\pi_V$ onto $R$:*

$$\pi_V^{\downarrow R} = (\otimes_{C \in \mathcal{C}} \pi_C)^{\downarrow R} = \pi_R \otimes \left(\otimes_{C \in \text{ne}(R)} \pi_{C \to R}\right).$$

### 3.6 COMPUTING OPTIMUM POLICIES BY MESSAGE PASSING

This section is concerned with showing how to find optimum policies for extremal decision nodes by message passing in the junction tree $\mathcal{T}$.

Let $\pi_W = (p_W, u_W)$ be a potential. The *contraction* of $\pi_W$, denoted cont$(\pi_W)$, is defined as the real function on $\mathcal{X}_W$ given by

$$\text{cont}(\pi_W) = p_W u_W.$$

Accordingly, for the joint potential $\pi_V$ defined by (4) we have

$$\text{cont}(\pi_V) = f(x)U(x). \qquad (5)$$

It is easily shown that for a potential $\pi_W$ on $W$ and $W_1 \subseteq W$ we have

$$\text{cont}(\pi_W^{\downarrow W_1}) = \sum_{W \setminus W_1} \text{cont}(\pi_W). \qquad (6)$$

To compute an optimum policy for an extremal decision node $d$, one first note that by (5) and (6)

$$\sum_{x_{V \setminus \text{fa}(d)}} f(x)U(x) = \text{cont}(\pi_V^{\downarrow \text{fa}(d)}).$$

Consequently, an optimum policy for $d$ can be found as follows. First one identifies a clique, say $R$, that contains fa$(d)$. The compilation of a LIMID $\mathcal{L}$ to a junction tree $\mathcal{T}$ guarantees the existence of such a clique. Then the following steps are carried out (cf. Corollary 1 and Theorem 1):

1. **Collect:** Collect to $R$ to obtain $\pi_R^* = \pi_V^{\downarrow R}$ as in Theorem 4.

2. **Marginalize:** Compute $\pi_{\text{fa}(d)}^* = (\pi_R^*)^{\downarrow \text{fa}(d)}$.

3. **Contract:** Compute the contraction $c_{\text{fa}(d)}$ of $\pi_{\text{fa}(d)}^*$.

4. **Optimize:** Define $\tilde{\delta}_d(x_{\text{pa}(d)})$ for all $x_{\text{pa}(d)}$ as the distribution degenerate at a point $x_d^*$ satisfying (cf. Corollary 1)

$$x_d^* = \arg \max_{x_d} c_{\text{fa}(d)}(x_d, x_{\text{pa}(d)}).$$

Note that all the computations apart from the second step are local in the root clique $R$.

Recall that, in SINGLE POLICY UPDATING, when an optimum policy $\tilde{\delta}_d$ for $d$ has been computed, $d$ is converted into a chance node with $\tilde{\delta}_d$ as its associated probability function. To make an equivalent conversion in our junction tree, we simply multiply $\tilde{\delta}_d$ onto the probability part of any clique containing fa$(d)$.

### 3.7 COMPUTING THE OPTIMAL STRATEGY BY PARTIAL COLLECT PROPAGATIONS

Suppose we have transformed a soluble LIMID $\mathcal{L}$ with exact solution ordering $d_1, \ldots, d_k$ into a junction tree $\mathcal{T}$. The propagation scheme presented here can be used to compute the optimum policies during SINGLE POLICY UPDATING.

As an initial step messages are collected towards any root clique $R_k$ which contains fa$(d_k)$. Then we compute an optimum policy for $d_k$, as described in the previous subsection, and the obtained policy is multiplied onto the probability part of $R_k$.

In a a similar manner the policy for $d_{k-1}$ can be computed: First, we identify a new root clique $R_{k-1}$ which contains fa$(d_{k-1})$. Then we could collect messages to $R_{k-1}$ as above; however, this usually involves a great deal of duplication. Instead we only need to pass messages along the (unique) path from the old root clique $R_k$ to $R_{k-1}$. This is done by first emptying the mailboxes on the path and then passing the messages. Note that after this 'partial' collection of messages, $R_{k-1}$ has received messages from all its neighbours. Now, an optimum policy for $d_{k-1}$ can be computed and the potential on $R_{k-1}$ is changed appropriately.

Proceeding in this way by successively collecting messages to cliques containing the families of the decisions we eventually compute all the optimum policies and thus the optimal strategy.

**Example 3** Fig. 8 shows how the propagation scheme works on our junction tree. For simplicity of exposition we have omitted the mailboxes in the junction tree.

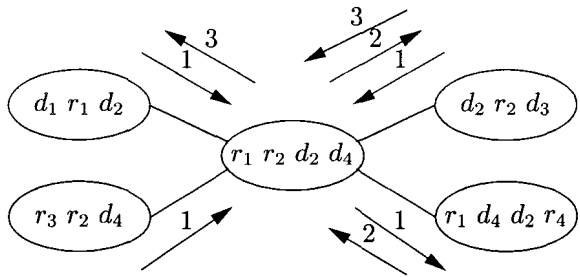

Figure 8: Passage of messages in the junction tree. The number attached to the arcs indicate the order that the messages are passed.

In our propagation scheme we successively collect messages towards cliques that contain the variables $\text{fa}(d_4), \ldots, \text{fa}(d_1)$ respectively. So we begin by collecting messages to clique $\{r_1, d_4, d_2, r_4\}$ since it contains $\text{fa}(d_4) = \{d_4, r_4, d_2\}$. Then we compute an optimum policy for $d_4$ and modify the probability part on the clique by multiplying it with the obtained policy for $d_4$. Now we partial collect messages towards clique $\{d_2, r_2, d_3\}$ because it contains $\text{fa}(d_3)$. After computing an optimum policy for $d_3$ and modifying the potential appropriately we partial collect messages towards clique $\{d_1, r_1, d_2\}$. Note that this clique not only contains $\text{fa}(d_2)$ but it also contains $\text{fa}(d_1)$, and thus we need not pass any more messages.

## 4  REFINEMENT OF THE ALGORITHM

Because multiple collect operations are performed in $\mathcal{T}$, we may pass many messages in the course of the evaluation of all the decisions. In the present section we give a condition for certain collect operations being unnecessary.

Suppose, at some stage in the algorithm, that the policy for decision $d$ is to be computed, and let $R$ be any clique containing $\text{fa}(d)$. In order to compute an optimum policy for $d$ we collect messages towards $R$. The following theorem states a condition for when a message from a neighbour of $R$ is superfluous.

**Theorem 5** *Let $C$ be a neighbour of clique $R$. Then, whenever $S = C \cap R \subseteq \text{pa}(d)$, the optimum policy for $d$ can be computed without the message from $C$.*

**Proof:** Let $\pi_R = (p_R, u_R)$ be the potential on $R$ after combining it with the messages from all its neighbours except $C$. Further, suppose $S = R \cap C \subseteq \text{pa}(d)$ and let $\pi_S = (p_S, u_S)$ be the message from $C$. We need to show that for computation of the optimum policy for $d$ as described in Section 3.6, the message $\pi_S$ is not needed.

Using (6) we have

$$\text{cont}\left((\pi_R \otimes \pi_S)^{\downarrow \text{fa}(d)}\right) = \sum_{R \setminus \text{fa}(d)} \text{cont}(\pi_R \otimes \pi_S)$$
$$= \sum_{R \setminus \text{fa}(d)} p_R p_S (u_R + u_S)$$

and

$$\text{cont}\left(\pi_R^{\downarrow \text{fa}(d)}\right) = \sum_{R \setminus \text{fa}(d)} p_R u_R.$$

Clearly, as $S \subseteq \text{pa}(d) \subseteq \text{fa}(d)$ we have that

$$\sum_{R \setminus \text{fa}(d)} p_R p_S u_R = c_0 \,\text{cont}\left(\pi_R^{\downarrow \text{fa}(d)}\right), \qquad (7)$$

where $c_0 = p_S \geq 0$ depends on $x_{\text{pa}(d)}$ only.

Since $f \propto f_{\bar{q}}$, where $f$ and $f_{\bar{q}}$ are given in (2) and (3) we have

$$\sum_{R \setminus \text{fa}(d)} p_R p_S u_S = u_S \sum_{V \setminus \text{fa}(d)} f$$
$$= c u_S \sum_{V \setminus \text{fa}(d)} f_{\bar{q}},$$

where $c$ is a constant. Because

$$f_{\bar{q}}(x_{\text{fa}(d)}) = f_{\bar{q}}(x_{\text{pa}(d)}) f_{\bar{q}}(x_d \mid x_{\text{pa}(d)}) = f_{\bar{q}}(x_{\text{pa}(d)}) / |\mathcal{X}_d|$$

is constant for fixed $x_{\text{pa}(d)}$, this yields

$$\sum_{R \setminus \text{fa}(d)} p_R p_S u_S = c_1, \qquad (8)$$

where $c_1$ depends on $x_{\text{pa}(d)}$ only.

Combining (7) and (8) now yields for fixed $x_{\text{pa}(d)}$

$$\text{cont}\left((\pi_R \otimes \pi_S)^{\downarrow \text{fa}(d)}\right) = c_0 \,\text{cont}\left(\pi_R^{\downarrow \text{fa}(d)}\right) + c_1,$$

where $c_0 \geq 0$, i.e. for each fixed $x_{\text{pa}(d)}$, the quantities to be optimized with and without the message from $R$ are linearly and positively related. This completes the proof. •

The following example shows an application of Theorem 5.

**Example 4** The ID displayed in Fig. 9 was introduced by Jensen *et al.* (1994). Fig. 10 shows the minimal reduction of the LIMID version of the ID and Fig. 11 shows the junction tree of the minimal reduction.

In order to compute the optimum policy for $d_4$ we collect flows towards clique $C_4$ since $C_4$ contains $\text{fa}(d_4)$.

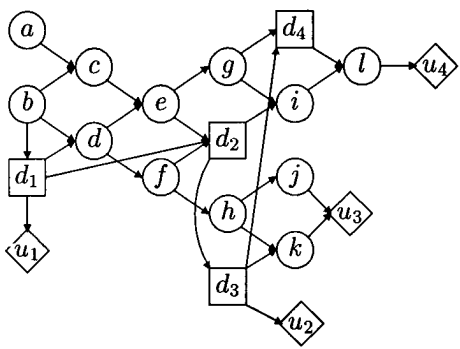

Figure 9: An ID

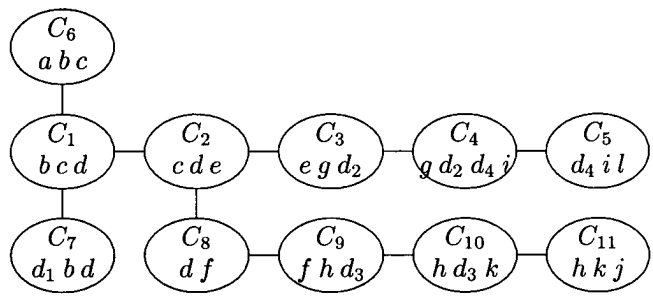

Figure 11: The junction tree for the LIMID in Fig. 10.

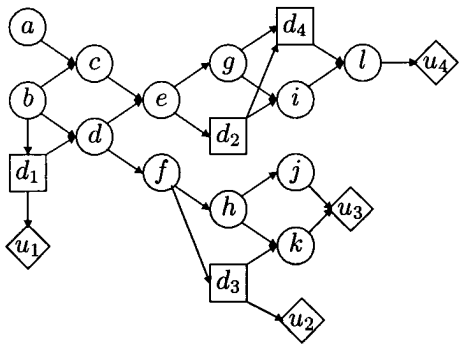

Figure 10: The minimal reduction of the LIMID version of the ID in Fig. 9.

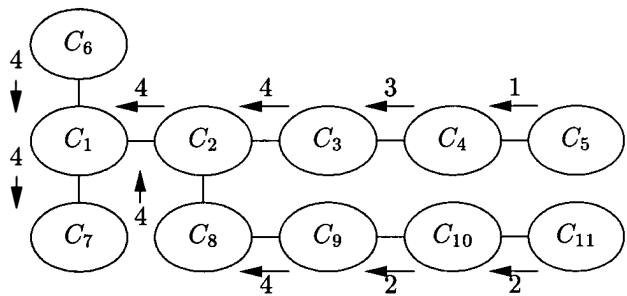

Figure 12: Flow of messages for the computation of the optimum policies using Theorem 5. The number attached to the arcs indicate the order that the messages are passed.

However, an application of Theorem 5 gives that the message from $C_3$ is unneccessary: $C_3 \cap C_4$ is a subset of pa($d_4$) in the minimal reduction in Fig. 10. Thus, we will only need the message from $C_5$. Furthermore, to compute the optimum policy for $d_3$ we collect flows towards clique $C_9$ since it contains fa($d_3$). But the flow from $C_8$ to $C_9$ is unnecessary because $C_8 \cap C_9$ is contained in pa($d_3$). Thus, only the flow from $C_{10}$ to $C_9$ is needed because it has not been computed earlier.

Continuing in this way, it turns out that only one flow along every edge in $\mathcal{T}$ is needed for the evaluation of the decision problem (see Fig. 12). So, by applying Theorem 5 we only need to pass 10 flows which is half the flows we would have passed using the partial propagation scheme presented earlier.

## 5 DISCUSSION

The method presented here transforms decision problems formulated as IDs into simpler structures, termed minimal LIMIDs, having the property that all requisite information for the computation of the optimal strategy is explicitly represented. It uses recursion to solve the decision problem by exploiting that the entire decision problem can be partitioned into a number of smaller decision problems each of which having one decision node only. A one-off process of compilation is then performed on the LIMID to produce a higher level graphical structure, the junction tree, that is particular well suited for efficient evaluation of each of the small decision problems.

The use of recursion is inspired by the well-known trick of Cooper (1988) and differs from methods in e.g. Shenoy (1992), and Jensen et al. (1994). By using recursion we do not require the storage of potentially large tables of intermediate results (see for instance Example 2).

As a consequence, our junction tree can always be made as small as the strong junction tree (Jensen et al. 1994), and in some cases our method can result in considerable reduction of evaluation time and memory. This reduction happens at two levels. At the first level, we obtain a smaller junction tree because we work in the reduced structure that only includes requisite information. At the second level, we obtain a smaller junction tree because we can triangulate the structure obtained without obeying order constraints.

On the other hand, our method typically passes more messages than the strong junction tree method, partly because our junction tree have more (and smaller) cliques, partly because we perform several collect propagations. As the size of the maximal clique most often is crucial for the efficiency of local computation algorithms, our algorithm should generally be fast com-pared to traditional algorithms.

There are many opportunities to refine and extend this research. In particular it should be possible to reduce the number of messages that are passed in the junction tree. We have presented one such condition for a message being redundant, but a deeper insight in the partial propagation algorithm may reveal other redundant computations, and improve the efficiency of the algorithm. The new method presented here also opens the possibility of evaluating large and computationally prohibitive decision problems by approximating them with soluble LIMIDs. Work regarding this issue is described in Lauritzen and Nilsson (1999) and is still in progress.

## Acknowledgements

This research was supported by DINA (Danish Informatics Network in the Agricultural Sciences), funded by the Danish Research Councils through their PIFT programme.

## References

Cooper, G. F. (1988). A method for using belief networks as influence diagrams. In *Proceedings of the 4th Workshop on Uncertainty in Artificial Intelligence*, pp. 55–63. Minneapolis, MN.

Cowell, R. G., Dawid, A. P., Lauritzen, S. L., and Spiegelhalter, D. J. (1999). *Probabilistic Networks and Expert Systems*. Springer-Verlag, New York.

Howard, R. and Matheson, J. (1981). Influence diagrams. In *The Principle and Applications of Decision Analysis*, (ed. R. Howard and J. Matheson), pp. 719–62. Strategic Decisions Group, Menlo Park, Calif.

Jensen, F., Jensen, F. V., and Dittmer, S. L. (1994). From influence diagrams to junction trees. In *Proceedings of the 10th Conference on Uncertainty in Artificial Intelligence*, (ed. R. L. de Mantaras and D. Poole), pp. 367–73. Morgan Kaufmann Publishers, San Francisco, CA.

Kjærulff, U. (1992). Optimal decomposition of probabilistic networks by simulated annealing. *Statistics and Computing*, **2**, 19–24.

Lauritzen, S. L. and Nilsson, D. (1999). LIMIDs of decision problems. Research Report R-99-2024, Dept. of Mathematical Sciences, Aalborg University. Submitted to *Management Science*.

Ndilikilikesha, P. (1994). Potential influence diagrams. *International Journal of Approximate Reasoning*, **10**, 251–85.

Nielsen, T. D. and Jensen, F. V. (1999). Welldefined decision scenarios. In *Proceedings of the 15th Annual Conference on Uncertainty in Artificial Intelligence*, (ed. K. Laskey and H. Prade), pp. 502–11. Morgan Kaufmann Publishers, San Francisco, CA.

Olmsted, S. (1983). *On Representing and Solving Decision Problems*. PhD thesis, Stanford University.

Pearl, J. (1986). Fusion, propagation and structuring in belief networks. *Artificial Intelligence*, **29**, 241–88.

Shachter, R. (1986). Evaluating influence diagrams. *Operations Research*, **34**, 871–82.

Shachter, R. (1998). Bayes-ball: The rational pasttime (for determining irrelevance and requisite information in belief networks and influence diagrams). In *Proceedings of the Fourteenth Annual Conference on Uncertainty in Artificial Intelligence (UAI-98)*, pp. 48–487. Morgan Kaufmann Publishers, San Francisco, CA.

Shachter, R. (1999). Efficient value of information computation. In *Proceedings of the 15th Annual Conference on Uncertainty in Artificial Intelligence*, (ed. K. Laskey and H. Prade), pp. 594–601. Morgan Kaufmann Publishers, San Francisco, CA.

Shachter, R. and Ndilikilikesha, P. (1993). Using influence diagrams for probabilistic inference and decision making. In *Proceedings of the Ninth Conference on Uncertainty in Artificial Intelligence*, (ed. D. Heckermann and A. Mamdani), pp. 276–83. Morgan Kaufmann, Stanford, California.

Shachter, R. and Peot, M. A. (1992). Decision making using probabilistic inference methods. In *Proceedings of the Eighth Annual Conference on Uncertainty in Artificial Intelligence (UAI-92)*, pp. 276–83. Morgan Kaufmann Publishers, San Francisco, CA.

Shenoy, P. P. (1992). Valuation-based systems for Bayesian decision analysis. *Operations Research*, **40**, 463–84.

Shenoy, P. P. and Shafer, G. R. (1990). Axioms for probability and belief-function propagation. In *Uncertainty in Artificial Intelligence IV*, (ed. R. D. Shachter, T. S. Levitt, L. N. Kanal, and J. F. Lemmer), pp. 169–98. North-Holland, Amsterdam.

Zhang, N. L. (1998). Probabilistic inference in influence diagrams. In *Proceedings of the Fourteenth Annual Conference on Uncertainty in Artificial Intelligence (UAI-98)*, pp. 514–22. Morgan Kaufmann Publishers, San Francisco, CA.